\documentclass[fleqn,10pt,nonatbib]{wlscirep}
\usepackage{lineno}
\usepackage[T1]{fontenc}
\usepackage[utf8]{inputenc}

\usepackage[maxbibnames=9, maxcitenames=2, backend=biber, natbib]{biblatex}

\usepackage[english]{babel}

\usepackage{csquotes}

\usepackage{amsmath}
\usepackage{amssymb}

\usepackage{isomath}

\usepackage{booktabs}
\usepackage{array}

\usepackage{algorithm, algpseudocode}

\usepackage{graphicx}
\graphicspath{{img/}}

\usepackage{pdfpages}

\usepackage{tikz}
\usetikzlibrary{positioning,shapes.geometric,fit}
\usepackage{pgfplots}
\pgfplotsset{compat=1.16}

\usepackage{siunitx}
\newcolumntype{U}{S[table-format=2.1(1), separate-uncertainty=true]}
\usepackage{multirow}

\newcommand{\mydataset}{MedIMeta}
\newcommand{\yep}{\checkmark}
\newcommand{\nope}{--}
\newcommand{\meh}{$\sim$}
\title{A comprehensive and easy-to-use multi-domain multi-task medical imaging meta-dataset}% (\mydataset)}

\author[1]{Stefano Woerner}
\author[1]{Arthur Jaques}
\author[1,2]{Christian F. Baumgartner}
\affil[1]{Cluster of Excellence \enquote{Machine Learning: New Perspectives for Science}, University of Tübingen, Germany}
\affil[2]{Faculty of Health Sciences and Medicine, University of Lucerne, Switzerland}

\begin{abstract}
	While the field of medical image analysis has undergone a transformative shift with the integration of machine learning techniques, the main challenge of these techniques is often the scarcity of large, diverse, and well-annotated datasets.
    Medical images vary in format, size, and other parameters and therefore require extensive preprocessing and standardization, for usage in machine learning.
    Addressing these challenges, we introduce the Medical Imaging Meta-Dataset (\mydataset), a novel multi-domain, multi-task meta-dataset.
    \mydataset\ contains 19 medical imaging datasets spanning 10 different domains and encompassing 54 distinct medical tasks, all of which are standardized to the same format and readily usable in PyTorch or other ML frameworks.
    We perform a technical validation of \mydataset, demonstrating its utility through fully supervised and cross-domain few-shot learning baselines.
\end{abstract}
\begin{document}

\flushbottom
\maketitle

\thispagestyle{empty}

\begin{figure}[h]
	\includegraphics[width=.139\textwidth]{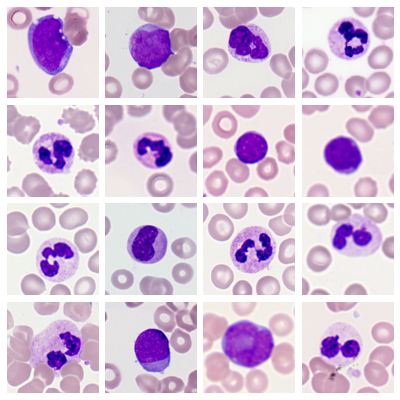}
	\includegraphics[width=.139\textwidth]{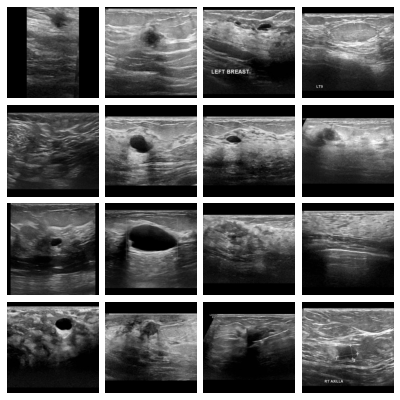}
	\includegraphics[width=.139\textwidth]{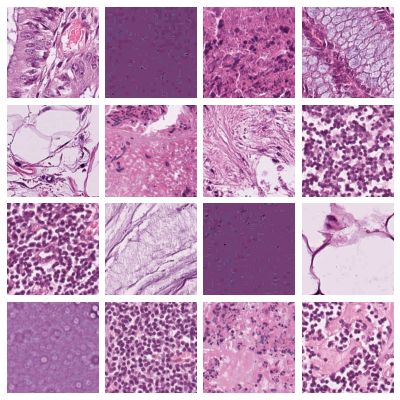}
	\includegraphics[width=.139\textwidth]{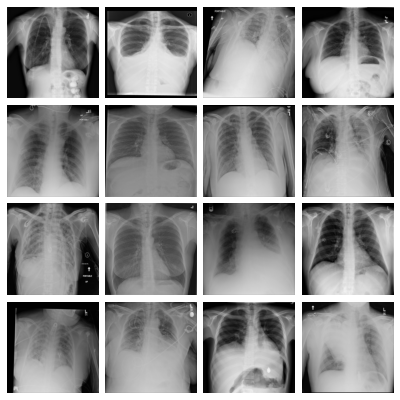}
	\includegraphics[width=.139\textwidth]{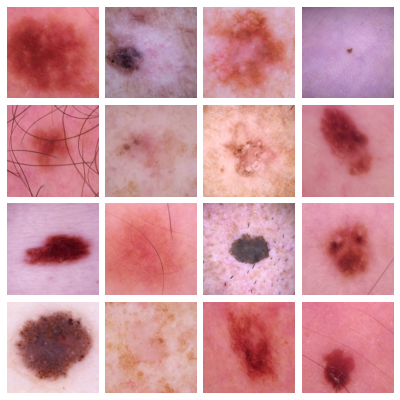}
	\includegraphics[width=.139\textwidth]{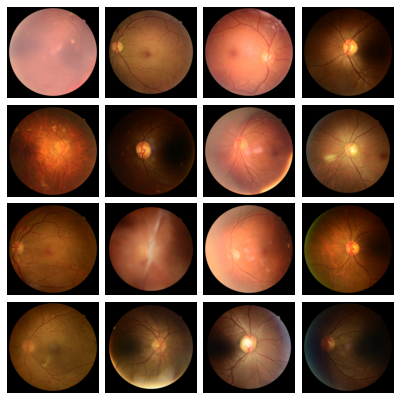}
	\includegraphics[width=.139\textwidth]{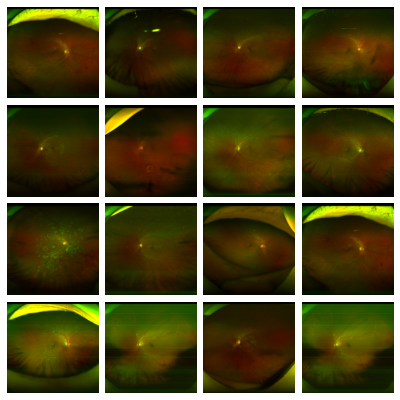}
	\includegraphics[width=.139\textwidth]{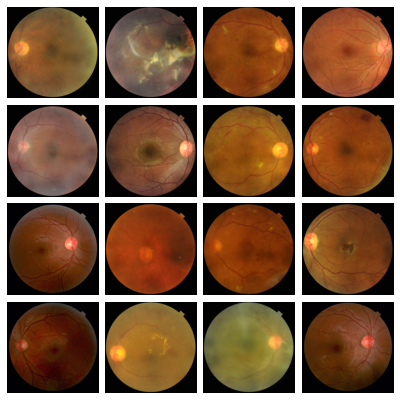}
	\includegraphics[width=.139\textwidth]{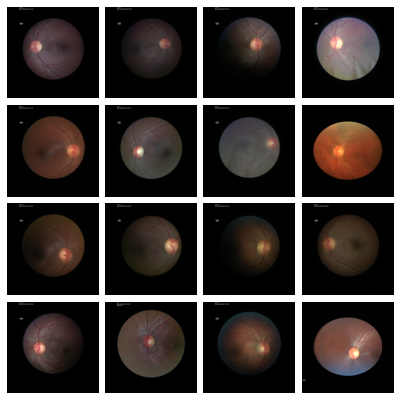}
	\includegraphics[width=.139\textwidth]{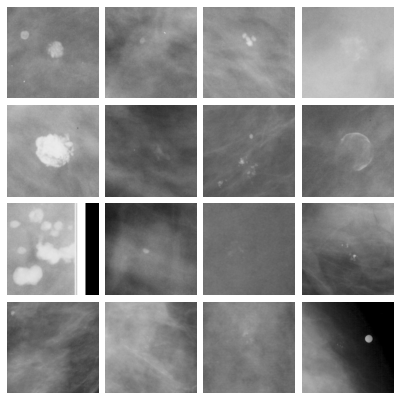}
	\includegraphics[width=.139\textwidth]{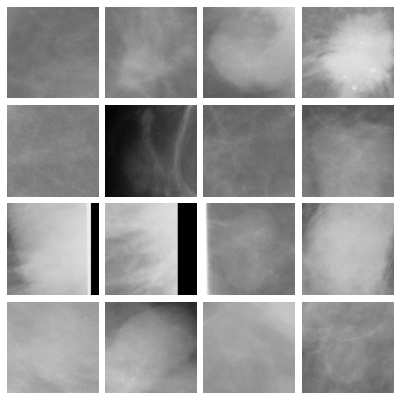}
	\includegraphics[width=.139\textwidth]{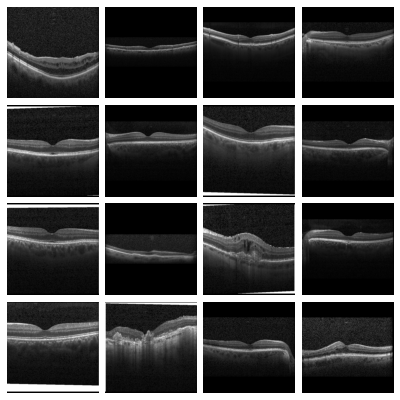}
	\includegraphics[width=.139\textwidth]{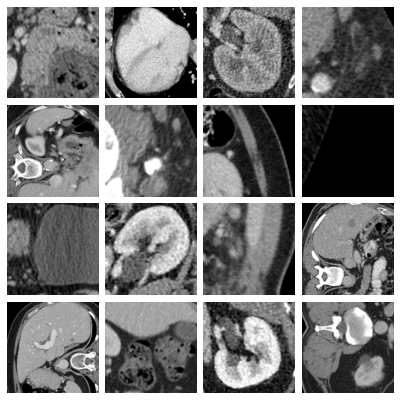}
	\includegraphics[width=.139\textwidth]{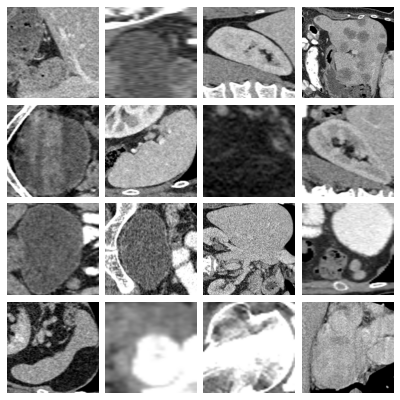}
	\includegraphics[width=.139\textwidth]{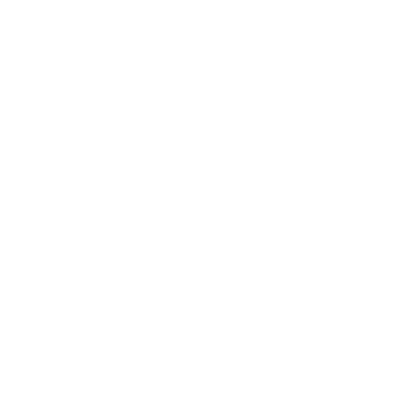}
	\includegraphics[width=.139\textwidth]{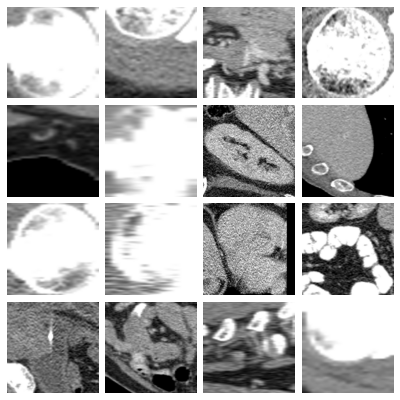}
	\includegraphics[width=.139\textwidth]{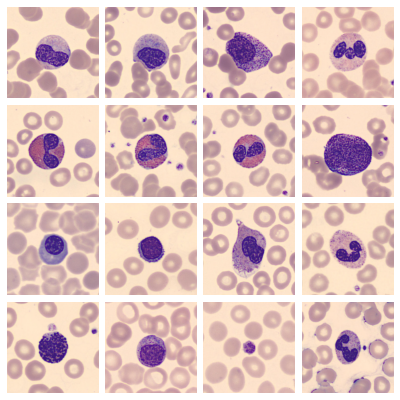}
	\includegraphics[width=.139\textwidth]{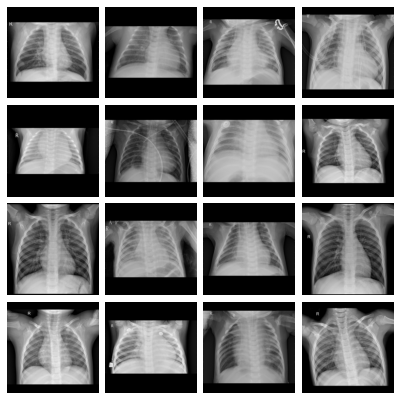}
	\includegraphics[width=.139\textwidth]{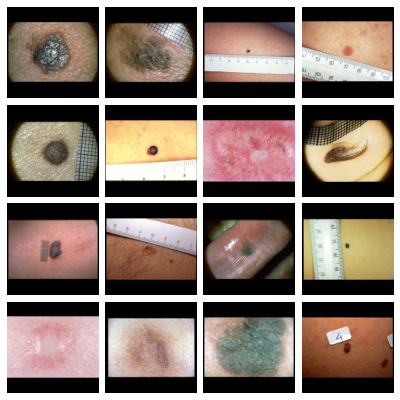}
	\includegraphics[width=.139\textwidth]{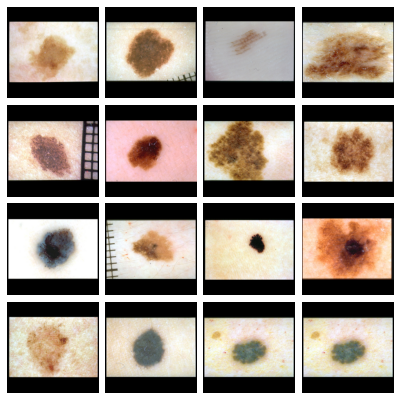}
	\includegraphics[width=.139\textwidth]{dataset-teasers/empty.png}
	\caption{Example images of all \mydataset\ datasets.}
    \label{fig:teaser}
\end{figure}

\section*{Background \& Summary}

In recent years, the field of medical image analysis has undergone a transformative shift with the integration of machine learning (ML) techniques fundamentally expanding the landscape of diagnostic and therapeutic strategies.
The advancement of this field hinges on the availability of large, diverse, and well-annotated datasets, which are pivotal for training robust and effective ML models.
However, the process of collecting raw medical images and bringing them to a format suitable for ML applications is often complex and fraught with challenges.
Medical images vary in format, size, and other parameters and therefore require extensive preprocessing and standardization, a task that becomes increasingly complex when working with multiple datasets from multiple domains that need to be integrated into a cohesive standardised format.

Another challenge is the scarcity of annotated datasets in medical imaging, particularly for rare diseases or specific conditions.
This limitation has sparked the exploration of few-shot learning (FSL) methodologies.
These try to learn from few examples and are designed to make predictions based on a limited number of training examples.
While humans are adept at learning and making accurate predictions from minimal information, achieving comparable levels of performance in machines remains a significant challenge.

Cross-domain few-shot learning (CD-FSL) allows models to leverage the information learned for a task in one domain and apply it to another task in another domain, potentially reducing the need for extensive data in every new task.
This ability is of immense value in these contexts, especially for rare conditions.
We define a domain as a particular type of subject, i.e.\ the studied anatomical region in the medical context, combined with a particular imaging modality.
Despite its potential, the cross-domain transfer poses significant challenges, given the inherent variability in imaging modalities, disease presentations, and data characteristics across different medical fields and clinics.
In addition to the variability in domains, the nature of tasks involved in medical image analysis varies considerably, encompassing a range of classification types such as binary, multi-class, or multi-label, and differing in the number of target labels or classes.
Developing algorithms capable of navigating these complexities is essential for effective knowledge transfer.
In order to facilitate the development of such algorithms, good preprocessing and standardization across the different domains are even more critical in this context.

Addressing these challenges, we introduce the Medical Imaging Meta-Dataset (\mydataset), a novel multi-domain, multi-task meta-dataset designed to facilitate the development and standardized evaluation of ML models and cross-domain FSL algorithms for medical image classification.
\mydataset\ contains 19 medical imaging datasets spanning 10 different domains and encompassing 54 distinct medical tasks, offering opportunities for both single-task and multi-task training.
Many of the tasks are diagnostic tasks or are tasks immediately relevant to a diagnosis.
Additionally, \mydataset\ contains auxiliary tasks (such as gender prediction) that may not have immediate clinical relevance but may nevertheless be of interest.
Furthermore, they have relevance for multi-task training or for training FSL algorithms which benefit from having a large amount of tasks.

For improved practicality, each dataset within the \mydataset\ dataset is standardized to a size of $224\times 224$ pixels which matches image size commonly used in pre-trained models.
Furthermore, the dataset comes with pre-made splits to ensure ease of use and standardized benchmarking.
We meticulously preprocessed the data and release a user-friendly Python package\footnote{\url{https://github.com/StefanoWoerner/medimeta-pytorch}}
%~\cite{mimeta-pytorch}
to directly load images for use in PyTorch.

This makes \mydataset\ exceptionally accessible to ML researchers.
This ease of access to a diverse set of realistic medical tasks with no need for additional preprocessing can serve as a bridge between medical professionals and the ML community, fostering interdisciplinary collaboration.
Moreover, \mydataset\ is an ideal platform for investigating cross-domain few-shot learning in medical imaging.
The rich array of tasks and domains presents an excellent opportunity to study and develop cross-domain few-shot learning techniques.

In addition to presenting the meta-dataset, this paper also presents a technical validation of \mydataset, demonstrating its utility through fully supervised and CD-FSL baselines.
The validation confirms the dataset's reliability and robustness, establishing it as a credible benchmark for research in ML for medical image analysis.

\subsection*{Related datasets}

Existing meta-datasets can be divided into two categories: those consisting of multiple datasets from a single domain, and those comprising data from multiple domains.
An overview is shown in Table~\ref{tab:dataset-comparison}.

\subsubsection*{Single-domain meta-datasets}

Single domain meta-datasets offer an easy way to benchmark few-shot learning techniques such as meta-learning.
One of the first meta-datasets in this category was Omniglot~\cite{lake2015human}, which consists of handwritten characters from a wide range of alphabets.
More challenging meta-datasets derived from natural images were obtained by subsampling the widely used ImageNet~\cite{ILSVRC15}, or CIFAR datasets~\cite{krizhevsky2009learning}.
Examples include the MiniImageNet~\cite{NIPS2016_90e13578} and the TieredImageNet~\cite{ren2018metalearning} datasets, as well as CIFAR-FS~\cite{bertinetto2018meta} and FC100~\cite{oreshkin2018tadam}.

\subsubsection*{Multi-domain meta-datasets}

While single-domain meta-datasets offer an easy standardized way to evaluate few-shot learning techniques they often lack realism.
In reality data are rarely from a single domain in FSL problems.
This realization has led to the release of several multi-domain meta-datasets.

The visual decathlon dataset~\cite{rebuffi2017learning} is one of the first multi-domain datasets.
It consists of 10 datasets from different visual tasks including traffic sign and flower recognition.
It also includes Omniglot~\cite{lake2015human} and CIFAR~\cite{krizhevsky2009learning} datasets discussed earlier.

Later \citeauthor{triantafillou2019meta} released another collection of 10 datasets coined the ``Meta-Dataset''~\cite{triantafillou2019meta}.
While it partially overlaps with the visual decathlon dataset, the ``Meta-Dataset'' was specifically designed to benchmark few-shot learning algorithms on multiple domains.
However, it does not contain any medical datasets.

With a similar motivation \citeauthor{vtab} released the Visual Task Adaptation Benchmark (VTAB)~\cite{vtab}.
This meta-dataset consists of 19 tasks, again, partially subsuming the previous datasets.
A particular property of the VTAB benchmark is the inclusion of datasets from three different domains covering natural image understanding, structured scene understanding, and specialized tasks.
The VTAB dataset is also to our knowledge the first meta-dataset to include medical images.

\citet{https://doi.org/10.48550/arxiv.2104.02638} later unified VTAB~\cite{vtab} and Meta-Dataset~\cite{triantafillou2019meta} into a larger Few-Shot Classification Benchmark.
\citet{https://doi.org/10.48550/arxiv.1912.07200} collected multiple previously available datasets from multiple domains for benchmarking CD-FSL methods.
Meta-Album~\cite{meta-album-2022} is a collection of 40 different dataset from multiple domains and follows a similar goal to our work.
While it contains more datasets than \mydataset, it does not contain any medical domains except for microscopy.

MedMNISTv2~\cite{yang2023medmnist} is a collection of 12 medical imaging datasets from 9 different domains.
It additionally contains 3D datasets.
It is similar in spirit to our work, but we go significantly beyond MedMNIST in the number of tasks and task realism.
The images in MedMNISTv2 have a very low resolution of $28\times 28$ pixels that obscures fine details that may be clinically relevant. 
In contrast, we process all images at high resolution and make them available with an image size of $224\times 224$ pixels, which allows to detect more clinically relevant features and is the typical resolution used in pretrained neural networks.
Additionally, MedMNISTv2 does not contain any multitask datasets.
In contrast, \mydataset contains a wide variety of tasks including binary, multi-class, and multi-label classification as well as ordinal regression.

In some of the overlapping datasets, we found significant problems with MedMNIST's preprocessing, which we improve upon.
Specifically, we found that some of the center-cropped images in MedMNIST had their relevant part, i.e., the part of the image showing the disease, cropped away.
We fix this problem by instead zero-padding these datasets.
Additionally, some datasets in MedMNISTv2 may have training-test splits that put images of the same subject in multiple splits.
We instead generated splits by taking subject information into account.

% ### original position of tab:dataset-comparison ###
\begin{table}
	\caption{Comparison of different datasets}
	\label{tab:dataset-comparison}
	{\centering
		\footnotesize
		\begin{tabular}{lllllllllllll}
			\toprule
			Name                   & \rotatebox{90}{\# domains} & \rotatebox{90}{\# datasets} & \rotatebox{90}{\# medical datasets} & \rotatebox{90}{task types *} & \rotatebox{90}{\# tasks} & \rotatebox{90}{\# images} & \rotatebox{90}{image size} & \rotatebox{90}{multi-domain} & \rotatebox{90}{different task types} & \rotatebox{90}{multi-task} & \rotatebox{90}{contains medical data} & \rotatebox{90}{uniform image size} \\ \midrule % & \rotatebox{90}{suitable image size} \\ \midrule
			Omniglot~\cite{lake2015human}                    & 1  & 1  & 0  & MC           & 1  & 32\,000      & $105\times 105$  & \nope & \nope & \nope & \nope & \yep  \\ % & \yep  \\
			\emph{mini}ImageNet~\cite{NIPS2016_90e13578}     & 1  & 1  & 0  & MC           & 1  & 60\,000      & $84\times 84$    & \nope & \nope & \nope & \nope & \yep  \\ % & \yep  \\
			\emph{tiered}ImageNet~\cite{ren2018metalearning} & 1  & 1  & 0  & MC           & 1  & 779\,165     & $84\times 84$ ** & \nope & \nope & \nope & \nope & \yep  \\ % & \yep  \\
			CIFAR-FS~\cite{bertinetto2018meta}               & 1  & 1  & 0  & MC           & 1  & 60\,000      & $32\times 32$    & \nope & \nope & \nope & \nope & \yep  \\ % & \yep  \\
			FC100~\cite{oreshkin2018tadam}                   & 1  & 1  & 0  & MC           & 1  & 60\,000      & $32\times 32$    & \nope & \nope & \nope & \nope & \yep  \\ % & \yep  \\
			Visual Decathlon~\cite{rebuffi2017learning}      & 1  & 10 & 0  & MC, (B)      & 10 & 1\,580\,558  & $\min(h,w)=72$   & \yep  & \nope & \nope & \nope & \meh  \\ % & \yep  \\
			Meta-Dataset~\cite{triantafillou2019meta}        & 7  & 10 & 0  & MC, (B)      & 10 & 53\,068\,000 &                  & \yep  & \nope & \nope & \nope & \nope \\ % &       \\
			VTAB~\cite{vtab}                                 & 3  & 19 & 1  & MC, (B)      & 19 & 2\,244\,000  &                  & \yep  & \nope & \nope & \meh  & \nope \\ % &       \\
			Meta-Album \emph{Mini}~\cite{meta-album-2022}    & 10 & 40 & 2  & MC           & 40 & 220\,950     & $128\times 128$  & \yep  & \nope & \nope & \meh  & \yep  \\ % & \yep  \\
			MedMNIST v2 (2D)~\cite{yang2023medmnist}         & 9  & 12 & 12 & MC, B, ML, O & 12 & 708\,069     & $28\times 28$    & \yep  & \yep  & \nope & \yep  & \yep  \\ % & \nope \\
			\midrule
			\textbf{\mydataset} (ours)                       & 10 & 19 & 19 & MC, B, ML, O & 54 & 399\,742     & $224\times 224$  & \yep  & \yep  & \yep  & \yep  & \yep  \\ % & \yep  \\
			\bottomrule\\
		\end{tabular}\\}
	\footnotesize
	* MC: multi-class, B: binary, ML: multi-label, O: ordinal.
	Brackets indicate that a task can be interpreted as that task type but is not defined as such in the respective benchmark dataset.\\
	** default settings.
\end{table}

While Meta-Dataset, VTAB, Meta-Album and MedMNIST contain different domains, only VTAB and MedMNIST contain a variety of different tasks.
\mydataset\ is the only data collection which contains datasets with multiple tasks, providing users of \mydataset\ with the option to train multi-task algorithms.
Out of these benchmarks, only
MedMNIST contains a significant number of \emph{medical} tasks.
Our proposed \mydataset\ contains 54 medical tasks and is easily extensible to tasks from other fields, like natural images.
Furthermore, we provide utilities for converting other datasets (e.g.\ ImageNet) into the \mydataset\ format allowing to easily integrate \mydataset\ with other data sources.

\section*{Methods}

We release the \mydataset\ dataset, a novel, highly standardized meta-dataset comprised of 19 publicly available datasets containing a total of 54 tasks.
In the following, we describe the source datasets and the data generation in detail. 

\subsection*{Dataset}

All datasets included in the \mydataset\ dataset have either been previously published under an open license that allows redistribution under a CC-BY-SA or CC-BY-SA-NC license, or we obtained an explicit permission to do so.
In addition to having an open license, we selected these datasets based on three criteria: suitability for defining a minimum of one classification task on the data, image size suitable for rescaling to our target size without producing noticeable artifacts, and a minimum of 100 images.
All images in \mydataset\ were standardized to an image size of $224\times 224$ pixels.
We provide pre-defined training, validation and testing splits for all 19 datasets.
If data splits were already defined in the source data, we used the pre-existing splits.
Otherwise we generated our own.
Most datasets include more than one classification task.
Typically, there is one main diagnostic task and several auxiliary tasks.
Most of these tasks were already present in the source datasets.
In some instances, we created additional tasks not present in the source data.
Table~\ref{tab:all-tasks} gives an overview of all dataset, tasks, and their key properties.
In the following, we describe in detail all datasets that \mydataset\ is comprised of using the \texttt{dataset\,ID}, as well as the full dataset name.
\begin{description}
    \item[\texttt{aml}, AML Cytomorphology:] Morphological dataset of leukocytes with expert-labeled single-cell images from peripheral blood smears of patients with acute myeloid leukemia (AML) and patients without signs of hematological malignancy, derived from the \emph{Munich AML Morphology Dataset}~\cite{matek_human-level_2019}.
    The 18,365 original images were resized to $224\times 224$ pixels using bi-cubic interpolation (no images were up-scaled) and converted to RGB format by removing the transparency channel.
    We adopted the original multi-class classification task with 15 morphological classes from the source dataset.
    
    \item[\texttt{bus}, Breast Ultrasound:] Dataset of breast ultrasound images of women between 25 and 75 years old, derived from the \emph{Breast Ultrasound Images Dataset}~\cite{al-dhabyani_dataset_2020}.
    The 780 original images were converted to grayscale and their masks to binary format.
    Images and masks were zero-padded to a square shape and resized to $224\times 224$ pixels using bi-cubic interpolation and nearest neighbor interpolation respectively (no images were up-scaled).
    The multi-class tumor classification task with normal, benign, and malignant examples was adopted without modifications from the source dataset.
    We additionally defined a binary classification task between malignant tumors and other images.
    
    \item[\texttt{crc}, Colorectal Cancer:] Dataset of image patches from hematoxylin \& eosin (H\&E) stained histological images of human colorectal cancer (CRC) and healthy tissue, derived from the \emph{NCT-CRC-HE-100K} and \emph{CRC-VAL-HE-7K} datasets~\cite{kather_100000_2018}.
    The 107,180 original images from the training and validation sets were not modified, as they already had the right shape and size for \mydataset.
    We adopted the multi-class tissue classification task with 9 labels from the source dataset without modifications.
    
    \item[\texttt{cxr}, Chest X-ray Multi-disease:] Dataset of frontal-view X-ray chest images, derived from the \emph{ChestX-ray14} dataset~\cite{wang_chestx-ray8_2017}.
    The 112,120 original images were resized to $224\times 224$ pixels using bi-cubic interpolation (no images were up-scaled); the 519 images that were originally in RGBA format were converted to grayscale.
    We provide a multi-label thorax disease classification task with 14 labels adopted without modifications from the source dataset.
    We additionally provide a binary classification task of the patient sex derived from the labels present in the original data.
    
    \item[\texttt{derm}, Dermatoscopy:] Dataset of dermatoscopic images of common pigmented skin lesions from different populations acquired and stored by different modalities, derived from the \emph{HAM10000} dataset~\cite{tschandl2018ham10000}.
    The 11,720 original images were center-cropped to a square shape and resized to $224\times 224$ pixels using bi-cubic interpolation (no images were up-scaled).
    We provide the multi-class disease classification task with 7 labels defined in the challenge hosted by the International Skin Imaging Collaboration (ISIC)~\cite{codella2019skin}.
    
    \item[\texttt{dr\_regular}, Diabetic Retinopathy (Regular Fundus):] Dataset of fundus images with diabetic retinopathy grades and image quality annotations, derived from the \emph{DeepDRiD} dataset~\cite{liu2022deepdrid}.
    The 2,000 original images were center-cropped to a square shape and resized to $224\times 224$ pixels using bi-cubic interpolation (no images were up-scaled).
    Following the annotations present in the original data, we provide 5 tasks: diabetic retinopathy grade (ordinal regression task with 5 labels), sufficient image quality for gradability (binary classification task), strength of artifact (ordinal regression task with 6 labels), image clarity (ordinal regression task with 5 labels), and field definition (ordinal regression task with 5 labels).
    
    \item[\texttt{dr\_uwf}, Diabetic Retinopathy (Ultra-widefield Fundus):] Dataset of ultra-widefield fundus images with annotations for diabetic retinopathy grading, derived from the \emph{DeepDRiD} dataset~\cite{liu2022deepdrid}.
    Only the 250 original images without missing labels were kept.
    They were center-cropped to a square shape and resized to $224\times 224$ pixels using bi-cubic interpolation (no images were up-scaled).
    We adopted the DR grading task (ordinal regression) with 5 labels from the source dataset without modifications.
    
    \item[\texttt{fundus}, Fundus Multi-disease:] Multi-disease retinal fundus dataset of images captured using three different fundus cameras with 45 conditions annotated through adjudicated consensus of two senior retinal experts as well as an overall disease presence label, derived from the \emph{Retinal Fundus Multi-disease Image Dataset}~\cite{pachade2021retinal}.
    The 3,200 images were center-cropped to a square shape and resized to $224\times 224$ pixels using bi-cubic interpolation (no images were up-scaled).
    The original disease presence binary classification task and disease multi-label classification task with 45 labels were directly derived from the annotations provided by the original dataset. 
    
    \item[\texttt{glaucoma}, Glaucoma-specific fundus images:] Glaucoma-specific Indian ethnicity retinal fundus dataset of images acquired using three devices, where five expert ophthalmologists provided annotations on whether the subject is suspect for glaucoma or not, derived from the \emph{Chákṣu} dataset~\cite{kumar2023chaksu}.
    The 1,345 original images and their masks were zero-padded to a square shape and resized to $224\times 224$ pixels using bi-cubic interpolation and nearest neighbor interpolation respectively (no images were up-scaled).
    We used the glaucoma suspect majority vote annotation to derive a binary classification task.
    
    \item[\texttt{mammo\_calc}, Mammography (Calcifications):] Dataset of cropped regions of interest (calcifications), derived from the \emph{Curated Breast Imaging Subset of the Digital Database for Screening Mammography} (CBIS-DDSM)~\cite{lee_curated_2017}.
    The 1,872 images were obtained by extending the regions of interest (bounding boxes) to a square shape with a minimum size of $224\times 224$ pixels, and extracting the resulting region crops from the full original images.
    The region crops were then resized to $224\times 224$ pixels using bi-cubic interpolation.
    From the annotations, 3 tasks were derived: pathology presence (binary classification task), calcification type (multi-label classification task with 14 labels), and calcification distribution (multi-label classification task with 5 labels).
    
    \item[\texttt{mammo\_mass}, Mammography (Masses):] Dataset of cropped regions of interest (masses) from CBIS-DDSM\@.
    The 1,696 images were preprocessed as described for \emph{Mammography (Calcifications)}.
    From the annotations, 3 tasks were derived: pathology presence (binary classification task), mass shape (multi-label classification task with 8 labels), and mass margins (multi-label classification task with 5 labels).
    
    \item[\texttt{oct}, OCT:] Dataset of validated Optical Coherence Tomography (OCT) images labeled for disease classification, derived from~\cite{kermany_identifying_2018}.
    The 84,484 original images were center-cropped to a square shape and resized to $224\times 224$ pixels using bi-cubic interpolation (no images were up-scaled).
    The original dataset contains a multi-class disease classification task, with three different diseases and a healthy class, which we adopt without modifications.
    Additionally, we provide a binary task for whether the image warrants urgent referral to a specialist based on the annotations present in the original data.
    
    \item[\texttt{organs\_axial}, Axial Organ Slices:] Dataset of axial image slices of 11 different organs, extracted from the \emph{Liver Tumor Segmentation Benchmark} (LiTS) dataset~\cite{bilic2023liver} and the corresponding organ bounding box annotations from~\cite{xu2019efficient}.
    We derived a multi-class organ classification task with 11 labels by extracting a cropped image of each individual organ in each of the CT volumes using the bounding box annotations.
    We obtained a total of 1,645 organ images images.
    We removed 106 images for which the voxel size information was missing.
    The axes on one image were permuted to bring it to the same format as the other images.
    The images and masks were sliced from the original 3D volumes by taking the center of the organ bounding box in the axial plane.
    The Hounsfield-Units of the images were transformed into grayscale images by applying a window with a width of 400 and a level of 50, which are typical values for abdominal CT imaging.
    The images and masks were cropped to a square size in the physical space, by centering at the center of the bounding box and expanding the smaller side.
    The resulting images and masks were resized to $224\times 224$ pixels using bi-cubic and nearest neighbor interpolation, respectively.
    For visualization purposes, we additionally provide images averaged over the 10\% central slices with the projected bounding boxes of all organs extracted from the image drawn on top.
    
    \item[\texttt{organs\_coronal}, Coronal Organ Slices:] Dataset of coronal image slices of 11 different organs, extracted from the LiTS dataset.
    The images were processed the same as described for the \emph{Axial Organ Slices} dataset, except that the coronal projections were used.
    
    \item[\texttt{organs\_sagittal}, Sagittal Organ Slices:] Dataset of sagittal image slices of 11 different organs, extracted from the LiTS dataset.
    The images were processed the same as described for the \emph{Axial Organ Slices} dataset, except that the sagittal projections were used.
    
    \item[\texttt{pbc}, Peripheral Blood Cells:] Dataset of microscopic peripheral blood cell images of individual normal cells, captured from individuals without infection, with hematologic or oncologic disease and free of any pharmacologic treatment at the moment of blood collection, derived from~\cite{acevedo_dataset_2020}.
    The 17,092 original images were center-cropped to a square shape and resized to $224\times 224$ pixels using bi-cubic interpolation (no images were up-scaled).
    We adopted the original multi-class blood cell classification task with 8 labels from the source dataset without modifications to the annotations.
    
    \item[\texttt{pneumonia}, Pediatric Pneumonia:] Dataset of pediatric chest X-ray images labeled for pneumonia classification, derived from~\cite{kermany_identifying_2018}.
    The 5,856 original images were zero-padded to a square shape and resized to $224\times 224$ pixels using bi-cubic interpolation (some images were up-scaled); the 283 images that were originally in RGB format were converted to grayscale.
    From the original annotations, we derived a binary classification task for pneumonia presence as well as a multi-class task differentiating between normal images, bacterial pneumonia and viral pneumonia.
    
    \item[\texttt{skinl\_derm}, Skin Lesion Evaluation (Dermoscopy):] A dataset containing dermoscopic color images of skin lesions, along with corresponding labels for seven different evaluation criteria and the diagnosis, derived from~\cite{derm7pt}.
    The images were zero-padded to obtain a square image and then resized to $224\times 224$ pixels using bi-cubic interpolation.
    We adopted an overall diagnostic multi-class task, as well as separate classification tasks for each of the seven diagnostic criteria from the source dataset.
    Tasks containing infrequent labels have additional grouped versions which bundle the infrequent labels together into more frequent labels.
    This grouping is provided by the source dataset.
    
    \item[\texttt{skinl\_photo}, Skin Lesion Evaluation (Clinical Photography):] A dataset containing clinical color photography images of skin lesions, along with corresponding labels for seven different evaluation criteria and the diagnosis, derived from~\cite{derm7pt}.
    This dataset contains the same subjects as Skin Lesion Evaluation (Dermoscopy) and images were preprocessed in the same manner.
    The tasks were also identical to Skin Lesion Evaluation (Dermoscopy).
\end{description}

% ### original position of tab:all-tasks ###
\begin{table}
	\caption{All \mydataset\ tasks.}
	\label{tab:all-tasks}
	{\centering
		\scriptsize
        \begin{tabular}{lllrrrllr}
        \toprule
			Dataset Name & \rotatebox{90}{License} & Domain & \rotatebox{90}{\# Train Images} & \rotatebox{90}{\# Val Images} & \rotatebox{90}{\# Test Images} & Task Name & Task Target & \rotatebox{90}{\# Labels} \\ \midrule
        AML Cytomorphology & BY-SA & Microscopy & 12\,855 & 1\,836 & 3\,674 & morphological class & multi-class classification & 15 \\
        &&&&&&&& \\
        \multirow[c]{2}{*}{Breast Ultrasound} & BY-SA & \multirow[c]{2}{*}{Breast ultrasound} & \multirow[c]{2}{*}{546} & \multirow[c]{2}{*}{78} & \multirow[c]{2}{*}{156} & case category & multi-class classification & 3 \\
         & &  &  &  &  & malignancy & binary classification & 2 \\
        &&&&&&&& \\
        Colorectal Cancer Histopathology & BY-SA & Histopathology & 85\,000 & 15\,000 & 7\,180 & tissue class & multi-class classification & 9 \\
        &&&&&&&& \\
        \multirow[c]{2}{*}{Chest X-ray Multi-disease} & \multirow[c]{2}{*}{BY-SA} & \multirow[c]{2}{*}{Chest X-ray} & \multirow[c]{2}{*}{73\,421} & \multirow[c]{2}{*}{13\,103} & \multirow[c]{2}{*}{25\,596} & disease labels & multi-label classification & 14 \\
         & &  &  &  &  & patient sex & binary classification & 2 \\
        &&&&&&&& \\
        Dermatoscopy & BY-SA & Dermatoscopy & 10\,015 & 193 & 1512 & disease category & multi-class classification & 7 \\
        &&&&&&&& \\
        \multirow[c]{5}{*}{\begin{tabular}{@{}l@{}}Diabetic Retinopathy\\(Regular Fundus)\\\end{tabular}} & \multirow[c]{5}{*}{BY-SA} & \multirow[c]{5}{*}{Retinal fundus} & \multirow[c]{5}{*}{1\,200} & \multirow[c]{5}{*}{400} & \multirow[c]{5}{*}{400} & DR level & ordinal regression & 5 \\
         &  &  &  &  &  & Overall quality & binary classification & 2 \\
         &  &  &  &  &  & Artifact & ordinal regression & 6 \\
         &  &  &  &  &  & Clarity & ordinal regression & 5 \\
         &  &  &  &  &  & Field definition & ordinal regression & 5 \\
        \multirow[c]{3}{*}{\begin{tabular}{@{}l@{}}Diabetic Retinopathy\\(Ultra-widefield Fundus)\\\end{tabular}}\\
         & BY-SA & Retinal fundus & 150 & 50 & 50 & DR level & ordinal regression & 5 \\
        &&&&&&&& \\
        \multirow[c]{2}{*}{Fundus Multi-disease} & \multirow[c]{2}{*}{BY-SA} & \multirow[c]{2}{*}{Retinal fundus} & \multirow[c]{2}{*}{1\,920} & \multirow[c]{2}{*}{640} & \multirow[c]{2}{*}{640} & disease presence & binary classification & 2 \\
         &  &  &  &  &  & disease labels & multi-label classification & 45 \\
        &&&&&&&& \\
        Glaucoma-specific fundus images &  & Retinal fundus & 857 & 152 & 336 & Glaucoma suspect & binary classification & 2 \\
        &&&&&&&& \\
        \multirow[c]{3}{*}{Mammography (Calcifications)} & \multirow[c]{3}{*}{BY-SA} & \multirow[c]{3}{*}{Mammography} & \multirow[c]{3}{*}{1\,332} & \multirow[c]{3}{*}{214} & \multirow[c]{3}{*}{326} & pathology & binary classification & 2 \\
         &  &  &  &  &  & calc type & multi-label classification & 14 \\
         &  &  &  &  &  & calc distribution & multi-label classification & 5 \\
        \\
        \multirow[c]{3}{*}{Mammography (Masses)} & \multirow[c]{3}{*}{BY-SA} & \multirow[c]{3}{*}{Mammography} & \multirow[c]{3}{*}{1\,126} & \multirow[c]{3}{*}{192} & \multirow[c]{3}{*}{378} & pathology & binary classification & 2 \\
         &  &  &  &  &  & mass shape & multi-label classification & 8 \\
         &  &  &  &  &  & mass margins & multi-label classification & 5 \\
        \\
        \multirow[c]{2}{*}{OCT} & \multirow[c]{2}{*}{BY-SA} & \multirow[c]{2}{*}{OCT} & \multirow[c]{2}{*}{91\,615} & \multirow[c]{2}{*}{16\,694} & \multirow[c]{2}{*}{1\,000} & disease class & multi-class classification & 4 \\
         &  &  &  &  &  & urgent referral & binary classification & 2 \\
        \\
        Axial Organ Slices & BY-NC-SA & Abdominal CT & 871 & 156 & 618 & organ label & multi-class classification & 11 \\
        \\
        Coronal Organ Slices & BY-NC-SA & Abdominal CT & 871 & 156 & 618 & organ label & multi-class classification & 11 \\
        \\
        Sagittal Organ Slices & BY-NC-SA & Abdominal CT & 871 & 156 & 618 & organ label & multi-class classification & 11 \\
        \\
        Peripheral Blood Cells & BY-SA & Microscopy & 11\,964 & 1\,709 & 3\,419 & cell class & multi-class classification & 8 \\
        \\
        \multirow[c]{2}{*}{Pediatric Pneumonia} & \multirow[c]{2}{*}{BY-SA} & \multirow[c]{2}{*}{Chest X-ray} & \multirow[c]{2}{*}{4\,415} & \multirow[c]{2}{*}{817} & \multirow[c]{2}{*}{624} & pneumonia presence & binary classification & 2 \\
         &  &  &  &  &  & disease class & multi-class classification & 3 \\
        \\
        \multirow[c]{12}{*}{\begin{tabular}{@{}l@{}}Skin Lesion Evaluation\\(Dermoscopy)\\\end{tabular}} & \multirow[c]{12}{*}{BY-SA} & \multirow[c]{12}{*}{Dermatoscopy} & \multirow[c]{12}{*}{413} & \multirow[c]{12}{*}{203} & \multirow[c]{12}{*}{395} & Diagnosis & multi-class classification & 15 \\
        &  &  &  &  &  & Diagnosis grouped & multi-class classification & 5 \\
        &  &  &  &  &  & Pigment Network & multi-class classification & 3 \\
        &  &  &  &  &  & Blue Whitish Veil & binary classification & 2 \\
        &  &  &  &  &  & Vascular Structures & multi-class classification & 8 \\
        &  &  &  &  &  & Vascular Structures grouped & multi-class classification & 3 \\
        &  &  &  &  &  & Pigmentation & multi-class classification & 5 \\
        &  &  &  &  &  & Pigmentation grouped & multi-class classification & 3 \\
        &  &  &  &  &  & Streaks & multi-class classification & 3 \\
        &  &  &  &  &  & Dots and Globules & multi-class classification & 3 \\
        &  &  &  &  &  & Regression Structures & multi-class classification & 4 \\
        &  &  &  &  &  & Regression Structures grouped & binary classification & 2 \\
        \\
        \multirow[c]{12}{*}{\begin{tabular}{@{}l@{}}Skin Lesion Evaluation\\(Clinical Photography)\\\end{tabular}} & \multirow[c]{12}{*}{BY-SA} & \multirow[c]{12}{*}{Clinical skin imaging} & \multirow[c]{12}{*}{413} & \multirow[c]{12}{*}{203} & \multirow[c]{12}{*}{395} & Diagnosis & multi-class classification & 15 \\
         &  &  &  &  &  & Diagnosis grouped & multi-class classification & 5 \\
         &  &  &  &  &  & Pigment Network & multi-class classification & 3 \\
         &  &  &  &  &  & Blue Whitish Veil & binary classification & 2 \\
         &  &  &  &  &  & Vascular Structures & multi-class classification & 8 \\
         &  &  &  &  &  & Vascular Structures grouped & multi-class classification & 3 \\
         &  &  &  &  &  & Pigmentation & multi-class classification & 5 \\
         &  &  &  &  &  & Pigmentation grouped & multi-class classification & 3 \\
         &  &  &  &  &  & Streaks & multi-class classification & 3 \\
         &  &  &  &  &  & Dots and Globules & multi-class classification & 3 \\
         &  &  &  &  &  & Regression Structures & multi-class classification & 4 \\
         &  &  &  &  &  & Regression Structures grouped & binary classification & 2 \\
        \bottomrule
        \end{tabular}\\
    }
\end{table}

We open-source all code for creating the above datasets from their respective source materials in \url{https://github.com/StefanoWoerner/medimeta-dataset-scripts}.
%~\cite{mimeta-dataset-scripts}.
Our published source code contains easy-to-use utility functions to extend \mydataset\ with additional datasets.
As examples, we provide several additional pipelines to create more datasets in the same format from other medical data which is publicly available but is not published under a license which allows redistribution of derivative work.

\subsection*{Python package}
We release a Python package called medimeta that enables data loading.
This package is installable via \texttt{pip install medimeta}.
Users can load data as single datasets or as cross-domain batches.
It also supports loading few-shot tasks with support and query sets.
Additionally, it is fully compatible with TorchCross~\cite{torchcross}, a library for cross-domain and few-shot learning.

\section*{Data Records}
We have organized and made available the data files of the \mydataset\ dataset on Zenodo in~\cite{mimeta-data}.
The data for all datasets within \mydataset\ can be accessed via the provided DOI. Each dataset is packaged in a single zip file for convenience and can be downloaded either separately or in one batch with all datasets.

Each zip file contains a structured and organized collection of data, designed to facilitate ease of use and comprehensive understanding of the dataset.
The content of these zip files is as follows:
\begin{description}
    \item[\texttt{images}] This folder contains all the image files for the dataset as uint8 TIFFs, named sequentially (e.g., \texttt{000000.tiff}, \texttt{000001.tiff}, etc.), ensuring easy access and exploration for human viewers.

    \item[\texttt{splits}] This directory includes text files (train.txt, val.txt, and test.txt) listing the image paths belonging to each respective split.

    \item[\texttt{original\_splits}] For source datasets with pre-existing split definitions, this directory contains those original splits, allowing users to adhere to the original data segmentation if desired.
    The format is the same as in the \texttt{splits} directory.

    \item[\texttt{task\_labels}] Each task within the dataset is accompanied by an \texttt{.npy} file named with the respective task's name (e.g., \texttt{task\_name\_1.npy}, \texttt{task\_name\_2.npy}, etc.) contained in this folder.
    Each \texttt{.npy} file contains a single NumPy array that represents the labels associated with that specific task.

    \item[\texttt{annotations.csv}] This file provides a comprehensive set of annotations, such as the patient id or imaging plane, for the images in the dataset, offering detailed insights and data points for those interested.
    It also contains all task labels to make them more readily accessible to human readers.

    \item[\texttt{images.hdf5}] This file contains all the images from the folder \texttt{images} formatted as a single dataset with dimensions $N \times C \times H \times W$, where N represents the number of images, C the number of channels, H the height, and W the width.
    The HDF5 file is useful for reading data in machine learning applications.

    \item[\texttt{info.yaml}] This file contains all relevant information about the dataset, including its ID, name and description, the number of images in each split, domain identifier, task definitions, and attribution information.

    \item[\texttt{LICENSE}] Each dataset is accompanied by its specific license file. We publish all datasets under a creative commons license. We published the majority of the datasets with the CC BY-SA 4.0 license\footnote{\url{https://creativecommons.org/licenses/by-sa/4.0/}}. However, some datasets are published with a non-commercial license\footnote{\url{https://creativecommons.org/licenses/by-nc-sa/4.0/}} due to the source material licensing. The specific CC license for each dataset is listed in table~\ref{tab:all-tasks}.
\end{description}

\section*{Technical Validation}

In order to validate our proposed dataset, we used it in two distinct learning scenarios.
First, we performed simple supervised training on each of the datasets on its primary tasks.
Secondly, we investigated the utility of our dataset for CD-FSL. In the following, we describe the experiment setups for the two scenarios in more detail.

\subsection*{Supervised learning experiments}
For the supervised experiments we trained ResNet-18 and ResNet-50 models~\cite{He_2016_CVPR} on the primary task for each dataset.
All networks were initialized with pre-trained weights from ImageNet~\cite{ILSVRC15}.
Early stopping was performed using the AUROC on the respective validation sets.
We performed a simple hyper-parameter search over data augmentation, learning rate and weight decay.
We note that the official test split of the Diabetic Retinopathy (Ultra-widefield Fundus) dataset does not contain any samples of the class \enquote{PDR}. Moreover, the dataset only contains two patients with this class in total, which prevents creating a custom train-test split with a better class balance. To account for this we trained and evaluated both models without this class.

\subsection*{Cross-domain few-shot learning (CD-FSL) experiments}

\begin{figure}
	\centering
	\includegraphics[width=.9\textwidth]{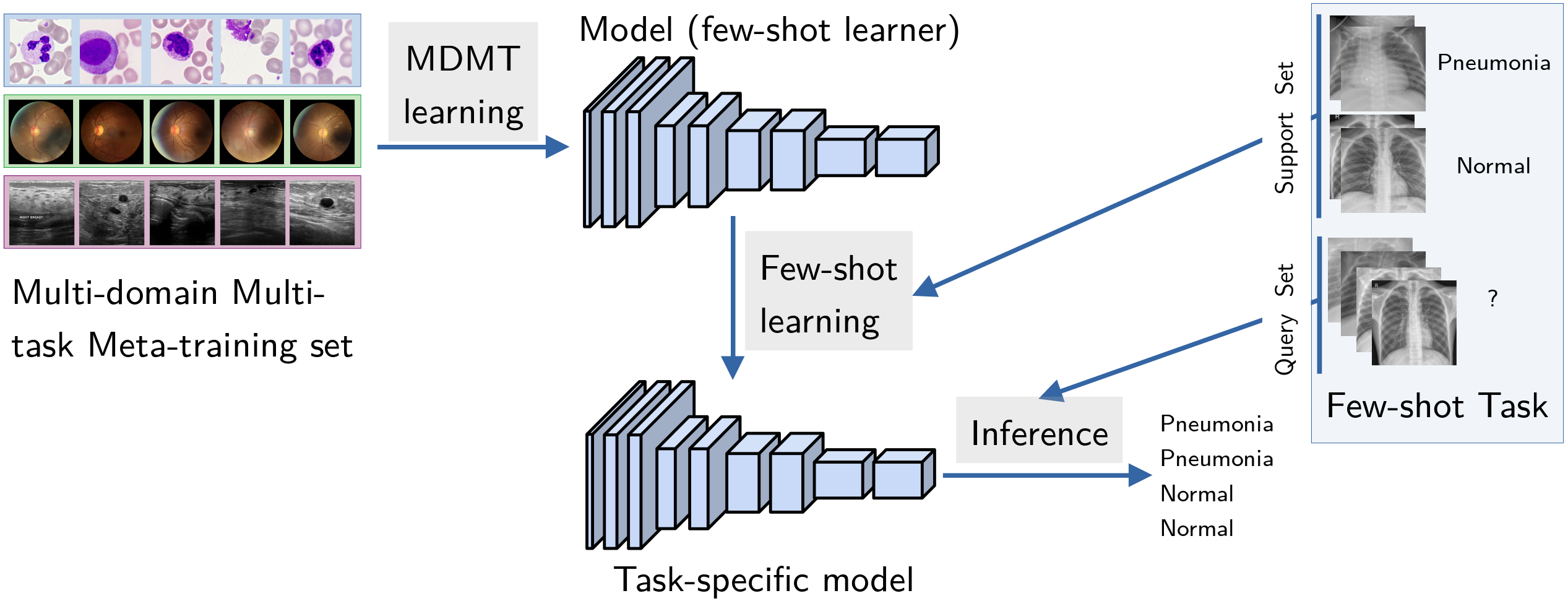}
	\caption{An overview of the CD-FSL scenario: The few-shot learner is first trained on the meta-dataset of highly diverse training data. It is then adapted to a new task from a new domain using the labeled examples from the support set of a few-shot task. Performance is assessed using a query set from the same task.}
	\label{fig:overview}
\end{figure}

In this evaluation, we compare 5-shot performance using three different CD-FSL approaches (described below).
We evaluated the performance for each dataset in a leave-one-out fashion using one task as target task, and the other tasks from all datasets with no domain overlap or subject overlap as source tasks to transfer knowledge from.
The knowledge transfer was achieved by first training a common backbone on the source tasks, and then fine-tuning the network to the target task using a small support set of 5 labeled examples per class.
Evaluation was performed on a distinct query set consisting of 10 samples from the same task.
We exclude all classes with less than 15 samples from the fine-tuning and evaluation, since at least 15 samples are needed to sample the distinct support and query sets.
Figure~\ref{fig:overview} illustrates the training and evaluation procedure.
Because performance can vary substantially between runs due to the quality of the 5 labeled examples, we reran the experiments 100 times for each target task to obtain a more robust estimation of the performance.

\subsubsection*{Baseline CD-FSL algorithms}
\label{sec:baseline-cd-fsl}

\paragraph{ImageNet pre-training (IM-PT):} The simplest baseline we investigated is simply initializing the common backbone network with ImageNet weights, and then directly fine-tuning to the target task. 

\paragraph{Multi-domain multi-task pre-training (mm-PT):} Pre-training using ImageNet lacks specificity to the medical domain.
Incrementally pre-training on a series of available datasets may offer a strategy to learn from many related datasets rather than just one~\cite{davidson2020sequential}.
However, incremental pre-training may suffer from catastrophic forgetting of earlier tasks~\cite{kirkpatrick2017overcoming}.
To address this issue, we propose a multi-domain multi-task pre-training schedule, where for each model update we sample a batch from a random source task.
This strategy may facilitate learning representations suitable for a wide range of tasks.
The algorithm is summarized in the supplemental materials.

\paragraph{Multi-domain Multi-task MAML (mm-MAML):} Model-agnostic meta-learning~\cite{maml} has been shown to be a promising strategy for CD-FSL~\cite{guo2020broader}.
MAML first ``learns to learn'' from a set of training tasks before learning the desired test task.
However, MAML assumes identical task types and number of classes for each task, which is not realistic in practical settings.
Here, we employed our previously proposed Multi-domain Multi-task MAML (mm-MAML) strategy~\cite{woerner2022strategies} where we used an individual classification layer for each class that is simply initialized with zeros.
The algorithm is summarized in the supplemental materials.

\subsection*{Results}
Table~\ref{tab:baseline-results} shows the results for the single-domain baselines.
It can be seen that the models were able to achieve a high performances in terms of AUROC for most datasets.
The comparatively lower performance on some of the tasks reflects the difficulty of these tasks.
All tasks with scores below 80 contain classes with very few examples, making these tasks more difficult to learn.
Since our fully supervised baselines are generic and simple methods not tailored to a specific task, we do not expect to see state-of-the-art performance on all datasets.
As expected, the more complex ResNet 50 model achieved slightly higher performance for most datasets compared to the ResNet 18.

\begin{table}[htbp]
	\caption{AUROC (\%) on the test set for the fully supervised baselines.}\label{tab:baseline-results}%
	\centering
	\scriptsize
	\begin{tabular}{@{}lSSSSSSSSSS@{}}
		\toprule
		{} &
		{aml} &
		{bus} &
		{crc} &
		{cxr} &
		{derm} &
		{\begin{tabular}{c}dr\_\\regular\\\end{tabular}} &
		{\begin{tabular}{c}dr\_\\uwf\\\end{tabular}} &
		{fundus} &
		{glaucoma} &
		{\begin{tabular}{c}mammo\_\\calc\\\end{tabular}} \\
		\midrule
		ResNet 18 & 99.3 & 97.0 & 99.1 & 79.3 & \bfseries 96.0 & 85.2 & 71.1 & 97.4 & \bfseries 83.8 & 77.6 \\
		ResNet 50 & \bfseries 99.4 & \bfseries 97.1 & \bfseries 99.5 & \bfseries 80.4 & 95.2 & \bfseries 85.9 & \bfseries 72.8 & \bfseries 97.7 & 83.6 & \bfseries 77.7 \\
		\bottomrule
	\end{tabular}\\
	\begin{tabular}{@{}lSSSSSSSSS@{}}
    	\toprule
    	{} &
    	{\begin{tabular}{c}mammo\_\\mass\\\end{tabular}}&
    	{oct}&
    	{\begin{tabular}{c}organs\_\\axial\\\end{tabular}}&
    	{\begin{tabular}{c}organs\_\\coronal\\\end{tabular}}&
    	{\begin{tabular}{c}organs\_\\sagittal\\\end{tabular}}&
    	{pbc}&
    	{pneumonia}&
    	{\begin{tabular}{c}skinl\_\\photo\\\end{tabular}}&
    	{\begin{tabular}{c}skinl\_\\derm\\\end{tabular}} \\
    	\midrule
    	ResNet 18 & 74.0 & 99.2 & \bfseries 99.8 & 99.1 & \bfseries 97.5 & \bfseries 100.0 & \bfseries 97.8 & \bfseries 79.5 & 89.0 \\
    	ResNet 50 & \bfseries 76.5 & \bfseries 99.7 & \bfseries 99.8 & \bfseries 99.2 & 97.4 & \bfseries 100.0 & 97.7 & 74.7 & \bfseries 90.4 \\
    	\bottomrule
	\end{tabular}\\
\end{table}%

\begin{table}[htbp]
	\caption{AUROC (\%) for the CD-FSL baselines averaged across 100 5-shot episodes using a ResNet18 and ResNet50. }\label{tab:few-shot-results}%
	\centering
	\scriptsize
    \begin{tabular}{@{}llUUUUUUUUUU@{}}
    \toprule
		{} & {} &
		{aml} &
		{bus} &
		{crc} &
		{cxr} &
		{derm} &
		{\begin{tabular}{c}dr\_\\regular\\\end{tabular}} &
		{\begin{tabular}{c}dr\_\\uwf\\\end{tabular}} &
		{fundus} &
		{glaucoma} &
		{\begin{tabular}{c}mammo\_\\calc\\\end{tabular}} \\
        {Training Type} & {Backbone} & {} & {} & {} & {} & {} & {} & {} & {} & {} & {} \\
        \midrule
        \multirow[c]{2}{*}{ImageNet} & ResNet18 & \bfseries 82.1(6) & 71.4(18) & 94.2(4) & \bfseries 58.5(6) & 74.4(10) & 69.4(11) & 53.0(16) & 75.1(26) & 55.0(29) & 61.4(26) \\
         & ResNet50 & 82.0(7) & \bfseries 72.9(17) & \bfseries 95.4(3) & 57.9(6) & \bfseries 74.9(8) & \bfseries 71.3(9) & \bfseries 54.0(16) & \bfseries 77.0(24) & 53.0(28) & \bfseries 64.2(28) \\
        \multirow[c]{2}{*}{MM-PFT} & \multirow[c]{1}{*}{ResNet18} & 77.9(8) & 66.2(17) & 85.8(6) & 55.5(5) & 69.7(8) & 65.1(11) & 51.2(15) & 64.2(29) & \bfseries 56.2(31) & 60.5(28) \\
         & \multirow[c]{1}{*}{ResNet50} & 79.6(7) & 69.7(17) & 89.7(5) & 57.1(6) & 72.1(10) & 67.9(10) & 52.3(16) & 69.8(25) & 53.4(26) & 60.1(31) \\
        \multirow[c]{1}{*}{MM-MAML} & \multirow[c]{1}{*}{ResNet18} & 70.3(7) & 61.5(18) & 82.4(6) & 54.4(5) & 68.6(8) & 63.3(11) & 51.1(15) & 58.4(28) & 52.2(28) & 58.3(28) \\
        \bottomrule
    \end{tabular}
    \\
    \begin{tabular}{@{}llUUUUUUUUU@{}}
        \toprule
        {} & {} &
    	{\begin{tabular}{c}mammo\\mass\\\end{tabular}}&
    	{oct}&
    	{\begin{tabular}{c}organs\_\\axial\\\end{tabular}}&
    	{\begin{tabular}{c}organs\_\\coronal\\\end{tabular}}&
    	{\begin{tabular}{c}organs\_\\sagittal\\\end{tabular}}&
    	{pbc}&
    	{pneumonia}&
    	{\begin{tabular}{c}skinl\_\\photo\\\end{tabular}}&
    	{\begin{tabular}{c}skinl\_\\derm\\\end{tabular}} \\
        {Training Type} & {Backbone} & {} & {} & {} & {} & {} & {} & {} & {} & {} \\
        \midrule
        \multirow[c]{2}{*}{ImageNet} & ResNet18 & 54.3(30) & 73.0(12) & \bfseries 96.5(3) & \bfseries 92.9(4) & \bfseries 89.0(5) & 89.6(6) & \bfseries 95.3(11) & 60.4(18) & 69.5(17) \\
         & ResNet50 & \bfseries 55.5(28) & \bfseries 74.9(13) & 94.6(3) & 88.8(4) & 85.6(5) & \bfseries 94.4(5) & 94.9(11) & 60.5(16) & \bfseries 70.8(18) \\
        \multirow[c]{2}{*}{MM-PFT} & \multirow[c]{1}{*}{ResNet18} & 55.4(29) & 68.9(13) & 92.1(3) & 87.0(4) & 84.8(5) & 85.0(8) & 88.5(18) & \bfseries 63.0(17) & 67.5(18) \\
         & \multirow[c]{1}{*}{ResNet50} & 54.2(32) & 67.0(14) & 91.3(4) & 86.9(5) & 85.9(5) & 88.5(8) & 91.4(16) & 59.7(17) & 65.4(18) \\
        \multirow[c]{1}{*}{MM-MAML} & \multirow[c]{1}{*}{ResNet18} & 53.6(28) & 63.6(12) & 88.6(4) & 82.4(5) & 81.6(5) & 83.7(7) & 90.9(14) & 56.6(15) & 67.5(17) \\
        \bottomrule
    \end{tabular}

\end{table}%

Table~\ref{tab:few-shot-results} shows 5-shot results for the CD-FSL baselines described earlier.
Surprisingly, simple fine-tuning from pre-trained ImageNet weights performed as well or better than fine-tuning from the two pre-fine-tuning and MAML baselines.
This can mean that the pre-fine-tuning method we chose is too simple to bring a meaningful benefit.
At the same time it is also apparent that simple methods for few-shot learning such as these do not achieve performance close to fully supervised training.
We therefore conclude that \mydataset\ offers high enough complexity for evaluating future few-shot methods, with the fully supervised results setting an upper bar for future few-shot methods to achieve.

\section*{Usage Notes}

All datasets contained in the \mydataset\ dataset~\cite{mimeta-data} can be downloaded from Zenodo.
Using the code provided in our data loaders repository\footnote{\url{https://github.com/StefanoWoerner/medimeta-pytorch}}%~\cite{mimeta-pytorch}
, all tasks in \mydataset\ can easily be loaded as PyTorch datasets for single-domain, cross-domain and few-shot scenarios.
No further pre-processing is required, but it is possible to provide any TorchVision transforms to the dataset class when initializing it.
A simple step-by-step example of how to load a single dataset is as follows.
\begin{enumerate}
    \item Download the zip file for the dataset you would like to use (e.g.\ OCT) from the Zenodo record~\cite{mimeta-data}, and extract it to a directory of choice.
    Let us use \texttt{./data/MedIMeta} here.
    \item \texttt{pip install medimeta}
    \item The data can now be used as a PyTorch dataset.
    The following code snippet would instantiate the dataset for the Disease task of the OCT dataset, assuming the data is stored in the \texttt{data/MedIMeta} directory.
    \begin{verbatim}
from medimeta import MedIMeta
dataset = MedIMeta("data/MedIMeta", "oct", "Disease")
    \end{verbatim}
\end{enumerate}
The repository also contains a directory \enquote{examples} which contains several usage examples for single-domain training, cross-domain training, and few-shot fine-tuning.

\subsection*{Practical limitations}
We have identified a number of practical limitations for usage of our dataset, which we briefly discuss in this section. 
Firstly, our meta-dataset contains several dataset which vary substantially in the number of samples. Some of these datasets are rather small, limiting their practical use cases for applications that require a large amount of data.
Some of the tasks are separated from their clinical context and therefore may lack clinical realism. For instance, medical professional holistically evaluate multiple mammography views of the same patient instead of only looking at a small region of interest. When balancing practicality for machine learning with clinical realism, we have consciously prioritized the former while keeping the clinical task as realistic as possible.
The scope of our meta-dataset does not completely encompass all modalities and anatomical regions typically seen in medical imaging and is comprised solely of 2D images. In clinical practice 3D images, videos, etc. are commonly used in addition to 2D images. Additionally clinicians often score images on a multidimensional scale, while most of the datasets included in \mydataset\ include only classification labels.
Another potential limitation is the common size and format of all images in \mydataset. This might not be optimal for every individual application domain.
However, this is a very significant advantage for ease of use and once again a conscious trade-off in favor of practicality for machine learning.

\section*{Code availability}
The algorithms, methodologies, and procedures used in this paper are fully documented in our accompanying code repositories.
All code has been developed in Python and builds on widely used libraries.
Specific dependencies and their versions are documented as requirements in the respective repositories, ensuring easy setup and reproducibility.

\begin{description}
    \item[\mydataset\ dataset scripts:] The source code for creating datasets from their respective materials contains all scripts needed for the creation of the 19 \mydataset\ datasets, utility functions for extending the \mydataset\ with additional datasets and all parameters used, such as image size.
    The code is available at \url{https://github.com/StefanoWoerner/medimeta-dataset-scripts}.
    %~\cite{mimeta-dataset-scripts}.
    
    \item[\mydataset\ data loaders:] Easy-to-use code for data loading from \mydataset\ is provided both for the single-domain and the cross-domain (few-shot) scenarios.
    The code is available at \url{https://github.com/StefanoWoerner/medimeta-pytorch}.
    %~\cite{mimeta-pytorch}.
    The package can be directly installed using \texttt{pip install medimeta}. %TODO make sure name is correct
    
    \item[Experiments:] The code for our experiments includes scripts for preprocessing, data augmentation, model training, evaluation, and other necessary utilities.
    It utilizes our data loader code described above.
    The code is available at \url{https://github.com/StefanoWoerner/medimeta-experiments}.
    %~\cite{mimeta-experiments}.
    
    \item[TorchCross library:] A general PyTorch few-shot learning and cross-domain learning library which is used in the \mydataset\ data loaders code.
    The code is available at \url{https://github.com/StefanoWoerner/torchcross}
    %~\cite{torchcross}
    and the package can be directly installed using \texttt{pip install torchcross}.
\end{description}

\printbibliography

%\bibliography{bibliography}

%\noindent LaTeX formats citations and references automatically using the bibliography records in your .bib file, which you can edit via the project menu. Use the cite command for an inline citation, e.g. \cite{Kaufman2020, Figueredo:2009dg, Babichev2002, behringer2014manipulating}. For data citations of datasets uploaded to e.g. \emph{figshare}, please use the \verb|howpublished| option in the bib entry to specify the platform and the link, as in the \verb|Hao:gidmaps:2014| example in the sample bibliography file. For journal articles, DOIs should be included for works in press that do not yet have volume or page numbers. For other journal articles, DOIs should be included uniformly for all articles or not at all. We recommend that you encode all DOIs in your bibtex database as full URLs, e.g. https://doi.org/10.1007/s12110-009-9068-2.

\section*{Acknowledgements}

Funded by the Deutsche Forschungsgemeinschaft (DFG, German Research Foundation) under Germany’s Excellence Strategy – EXC number 2064/1 – Project number 390727645.
The authors thank the International Max Planck Research School for Intelligent Systems (IMPRS-IS) for supporting Stefano Woerner.

\section*{Author contributions statement}
SW and CB conceived the dataset and the experiments, SW and AJ pre-processed and prepared the data, SW conducted the experiments.
All authors reviewed the manuscript.

\section*{Competing interests} %(mandatory statement)

The authors have no competing interests to declare. 

\end{document}

\flushbottom

\thispagestyle{empty}%
\makeatletter
\vskip-36pt%
{\raggedright\sffamily\bfseries\fontsize{20}{25}\selectfont \@title\par}%
\vskip10pt
{\raggedright\sffamily\fontsize{12}{16}\selectfont  \@author\par}
\vskip25pt
\makeatother

\thispagestyle{empty}

\section*{Training Algorithms}

\subsection*{Multi-domain multi-task pre-training}
To address the issue of catastrophic forgetting, we propose a multi-domain multi-task pre-training schedule (which we abbreviate with mm-PT), where the model is simultaneously pre-trained on all datasets in our meta-dataset.
This means that the model will be simultaneously trained on multiple heterogeneous domains and tasks.
To achieve this, the network is divided into a backbone and a classification head, where the last linear layer of the network is removed to create the backbone, and a linear layer is initialized as the head for each task.
The algorithm for mm-PT involves randomly sampling batches from each of the tasks and training the model on these batches. The appropriate head and loss function is used for each batch, as illustrated in algorithm~\ref{alg:mm-pt}.
When fine-tuning to the target task, a new head appropriate for the target task is attached to the backbone.

\begin{algorithm}[htbp]
	\caption{Multi-domain Multi-task Pre-training}\label{alg:mm-pt}
	\begin{algorithmic}[1] % The number specifies where the line numbering starts
		\Require
		Set $T$ of tasks,
		Datasets $D_t$ for tasks $t \in T$,
		Model $f$ with parameters $\theta$,
		Classification heads $l_t$ for $t \in T$
		
		\While{training has not converged}
		\State $t \gets$ Sample a task $t$ from $T$ with probability $\frac{|D_t|}{\sum_{j \in T}|D_j|}$
		\State $B \gets$ Sample a batch from dataset $D_t$
		\State $\theta \gets \theta - \nabla_{\theta} \mathcal{L}_t(l_t(f(B; \theta)))$
		\EndWhile
		\State \Return $\theta$
	\end{algorithmic}
\end{algorithm}

\subsection*{Multi-domain multi-task meta-learning}
The appearance of medical images can vary significantly depending on the imaging modality used.
Furthermore, the nature of tasks involved in medical image analysis varies considerably, encompassing a range of classifications such as binary, multi-class, or multi-label, and differing in the number of target labels or classes required.

To enable meta-learning in a multi-domain multi-task setting, we propose a simple extension of MAML~\cite{maml} that can learn from a diverse set of tasks prior to fine-tuning on the target task.
Although MAML can be applied to almost any neural network architecture, the assumption of a constant architecture becomes an issue when training with a set of diverse tasks: if tasks have diverse targets, at least the output layer has to change.
We have investigated attaching new per-task classification layers as an extension of MAML and how to initialize them in~\cite{woerner2022strategies}. Initialising the weights of the classification layer with zeros mitigates the performance drop experienced with the default Kaiming initialisation.
We therefore use this strategy as our default initialization scheme for the task-specific classification heads.
Additionally, in each episode we use the loss function which is appropriate for the respective task.

The proposed Multi-domain Multi-task MAML (mm-MAML) is trained simultaneously on all datasets in our meta-dataset by sampling few-shot tasks at random from each of the tasks and using the appropriate head and loss function.
At meta-test time, i.e. when fine-tuning to the target task, a new head and loss function is simply created for each target task, exactly as described in the previous section on transfer learning.
The training algorithm is described in Algorithm~\ref{alg:mm-maml}.

\begin{algorithm}[htbp]
	\caption{Multi-Domain Multi-Task MAML}\label{alg:mm-maml}
	\begin{algorithmic}[1] % The number specifies where the line numbering starts
		\Require
		Set $T$ of tasks,
		Datasets $D_t$ for tasks $t \in T$,
		Model $f$ with parameters $\theta$ and meta-parameters $\varphi$,
		The number of inner steps $I$
		
		\While{training has not converged}
		\State $t \gets$ Sample a task $t$ from $T$ % with probability $\frac{|D_t|}{\sum_{j \in T}|D_j|}$
		\State $C \gets$ Sample a set $C$ of classes from $\mathcal{P}(C_t)$
		\State $l \gets$ create a classification head $l$ for the task $t$ with classes $C$
		\State $S \gets$ Sample a support set from dataset $D_t$
		\State $Q \gets$ Sample a query set from dataset $D_t$
		\State $\theta \gets \varphi$
		\For{$i\in\{1\dots I\} $}
		\State $\theta \gets \theta - \nabla_{\theta} \mathcal{L}_t(l(f(S; \theta)))$
		\EndFor
		\State $\varphi \gets \varphi - \nabla_{\varphi} \mathcal{L}_t(l(f(Q; \theta)))$
		\EndWhile
		\State \Return $\theta$
	\end{algorithmic}
\end{algorithm}